\documentclass{article}

\usepackage[dblblindworkshop, final]{neurips_2026}
\workshoptitle{Sim2Science: ML with Imperfect Scientific Models}
\usepackage{amsmath}
\usepackage{soul}
\usepackage{textcomp}

\usepackage[utf8]{inputenc} 
\usepackage[T1]{fontenc}    
\usepackage[hidelinks]{hyperref}       
\usepackage{url}            
\usepackage{booktabs}       
\usepackage{amsfonts}       
\usepackage{nicefrac}       
\usepackage{microtype}      
\usepackage{xcolor}         
\usepackage{graphicx}       
\usepackage{subcaption}     
\usepackage{glossaries}
\usepackage{wrapfig}
\glsdisablehyper
\newacronym{atdm}{ATDM}{Advanced Theoretical Detector Model}
\newacronym{sbi}{SBI}{Simulation-Based Inference}
\newacronym{czt}{3D CZT DSD}{3D Cadmium Zinc Telluride Drift Strip Detector}
\newacronym{fwhm}{FWHM}{Full-Width-at-Half-Maximum}
\newacronym{mlp}{MLP}{Multi-Layer Perceptron}
\newacronym{cie}{CIE}{Charge Induction Efficiency}
\newacronym{adc}{ADC}{Analog-to-Digital Converter}
\newacronym{pde}{PDE}{Partial Differential Equation}
\newacronym{mev}{MeV}{megaelectronvolt}
\newacronym{psa}{PSA}{Pulse-Shape Analysis}

\title{Uncertainty-Aware 3D Position Reconstruction in CZT Drift-Strip Detectors}
\workshoptitle{Sim2Science}

\author{%
    Alejandro Valverde Mahou\thanks{DTU Compute, Technical University of Denmark, Kgs. Lyngby, Denmark.}\\
    \texttt{avama@dtu.dk}
    \And
    Michał Kossakowski\thanks{DTU Space, Technical University of Denmark, Kgs. Lyngby, Denmark.}\\
    \texttt{mmako@dtu.dk}
    \And
    Selina R. H. Owe\footnotemark[2]\\
    \texttt{shoowe@dtu.dk}
    \And
    Rikke S. Klausen\footnotemark[2]\\
    \texttt{rstkl@dtu.dk}
    \And
    Irfan Kuvvetli\footnotemark[2]\\
    \texttt{irfan@space.dtu.dk}
    \And
    Christian A. Naesseth\thanks{Informatics Institute, University of Amsterdam, Amsterdam, Netherlands.}\\
    \texttt{c.a.naesseth@uva.nl}
    \And
    Søren Hauberg\footnotemark[1]\\
    \texttt{sohau@dtu.dk}
}

\begin{document}
\maketitle
\begin{abstract}
Accurate three-dimensional interaction-position reconstruction is important for CZT drift-strip detectors, but conventional methods can suffer from degraded performance in challenging detector regions, while learned models provide limited information about prediction reliability. We investigate two uncertainty-aware approaches: deep ensembles of point-estimation networks and simulation-based inference (SBI), which estimates a posterior distribution over interaction position. Both approaches are trained on physics-based simulations and evaluated on both synthetic and experimental data. On synthetic data, uncertainty from both methods correlates with positioning error, with SBI showing a substantially stronger relationship. On experimental measurements, the deep ensemble achieves improved $z$-position resolution, while SBI provides more informative uncertainty estimates that highlight regions of reduced positional information and discrepancies between simulation and the physical detector. These results demonstrate the complementary strengths of ensemble-based reconstruction and posterior inference for uncertainty-aware detector position reconstruction.
\end{abstract}

\section{Introduction}

High-energy astrophysics and medical imaging rely on inferring particle interaction properties from detector signals, an inherently ill-posed inverse problem. In the \gls{mev} range, this is particularly challenging due to complex interaction physics, charge transport, and multi-channel signal formation. \gls{czt} detectors are promising for \gls{mev} gamma-ray detection, with applications ranging from space instrumentation to medical imaging \citep{van1998novel, van1998cdznte, owe2022evaluation}. In particular, 3D drift-strip architectures provide fine-grained interaction-position information \citep{kuvvetli2005pixelated, kuvvetli2024advancements}.

Position reconstruction is conventionally performed using  \gls{psa}, which can achieve submillimeter resolution \citep{budtz2017new}. However, reconstruction quality can deteriorate near detector boundaries, producing dead zones and regions of degraded resolution. Deep learning alternatives map from detector waveforms to interaction position. Such models require simulated training data and can achieve accurate point estimates, but provide no direct measure of prediction reliability. Furthermore, discrepancies between the physics-based simulator and the real detector, such as the point-charge approximation of the charge cloud used here, can introduce a sim-to-real gap that deterministic regression cannot quantify \citep{owe2026atdm}.

Uncertainty-aware methods offer a complementary approach. Ensembles quantify variability across independently trained predictors, while \gls{sbi} estimates a posterior distribution over interaction position conditioned on the measured signal and simulation model \citep{cranmer2020sbi}. We compare these approaches for \gls{czt} interaction-position reconstruction, examining how their uncertainty estimates relate to localization error and known detector anomalies. We further compare their reconstruction resolution with conventional \gls{psa} on experimental data and investigate whether \gls{sbi} uncertainty reveals discrepancies between simulation and the physical detector.\looseness=-1

\section{Detector and Data}
\label{sec:detector_data}

\textbf{3D Cadmium Zinc Telluride Drift Strip Detector.~~}
%
We consider a \gls{czt} drift-strip detector with dimensions $40\times5\times40,\text{mm}^3$, detailed by \citet{owe2022evaluation}. The detector has 69 readout channels: 24 anodes, 25 drift strips, and 20 cathodes. Each interaction produces a multi-channel pulse waveform from which the three-dimensional interaction position $(x,y,z)$ is reconstructed. 

Interaction positions can be reconstructed using conventional pulse-shape analysis. We use the algorithm of \citet{budtz2017new}, with the implementation applied in \citet{hauptmann2024evaluation}, as the conventional baseline for comparison. The method combines features of the anode, drift-strip, and cathode signals to estimate the three spatial coordinates.


\textbf{Training data} for the neural-network and \gls{sbi} approaches are generated using the \gls{atdm} \citep{owe2026atdm}, a physics-based simulation framework that combines electrostatic and \gls{cie} modelling with signal generation and electronics simulation. Because full-volume \gls{cie} simulation for all 69 channels is computationally prohibitive, the detector is represented by a set of precomputed sub-volumes covering the different electrode regions. These sub-volumes are translated across the detector geometry to generate pulse shapes for arbitrary interaction positions. The simulated signals are subsequently passed through the electronics model and converted from CIE to ADC counts. See Appendix~\ref{sec:atdm} for details.

The resulting synthetic events have the same multi-channel waveform format as the experimental data, with each event associated with a known three-dimensional interaction position. This enables supervised training of the neural networks and simulation-based inference.



\textbf{Experimental validation} (hereafter, real data) used the same $40\times5\times40,\mathrm{mm}^3$ \gls{czt} module modelled in the \gls{atdm} (Detector~2, \citep{hauptmann2024evaluation, owe2026atdm}). The detector was irradiated with a $^{137}$Cs source at 662~keV, matching the simulation energy. Three geometries were used: $0.2$~mm slit beams along $x$ and $z$ to quantify position resolution via the \gls{fwhm}, a $\sim45^\circ$ diagonal slit beam to assess localized reconstruction artifacts, and full illumination to assess global reconstruction behavior.

Events were filtered to isolate single-site interactions, and all output channels were baseline-normalized for consistency. No electrode-response calibration, such as ADC-to-keV conversion or non-linearity correction, was applied, preserving the raw pulse-shapes for the sim-to-real comparison.\looseness=-1

\section{Methods}

We compare two approaches to uncertainty-aware interaction-position reconstruction: a deep ensemble of point-estimation networks and \gls{sbi}. Both approaches share the same waveform encoder, providing a common signal representation for a direct comparison.

\textbf{Model architecture.~~}
Each event consists of 69 detector waveforms with 500 samples each. A shared one-dimensional convolutional encoder processes each waveform independently and maps it to a 16-dimensional embedding. The 69 embeddings are concatenated and projected to a 32-dimensional latent representation. The resulting representation is passed to either an ensemble member or an \gls{sbi} inference head. Full architectural details are given in Appendix~\ref{app:model_architecture}.

The encoder is pretrained as a \emph{denoising autoencoder}: Gaussian noise is added to the input waveforms, while the reconstruction target is the corresponding clean waveform. This encourages the latent representation to retain relevant signal information while being robust to measurement noise. After pretraining, the decoder is discarded and the encoder is used for both approaches.

\textbf{Deep Ensemble.~~}
Each ensemble member maps an input waveform $\mathbf{s}$ to a three-dimensional interaction position $\hat{\mathbf{x}}_m=f_{\theta_m}(\mathbf{s})$. We train $M=10$ independently initialized models with different data shuffling. The ensemble prediction is the mean $\bar{\mathbf{x}}=M^{-1}\sum_m\hat{\mathbf{x}}_m$, while epistemic uncertainty is estimated from the sample covariance $\mathbf{\Sigma}_{\mathrm{ens}}=(M-1)^{-1}\sum_m(\hat{\mathbf{x}}_m-\bar{\mathbf{x}})(\hat{\mathbf{x}}_m-\bar{\mathbf{x}})^\top$.

\textbf{Simulation-based inference.~~}
For \gls{sbi}, we model the conditional position distribution as a diagonal Gaussian, $p(\mathbf{x}\mid\mathbf{s})=\mathcal{N}(\mathbf{x};\boldsymbol{\mu}(\mathbf{s}),\operatorname{diag}(\boldsymbol{\sigma}^2(\mathbf{s})))$. The inference head predicts $(\boldsymbol{\mu},\log\boldsymbol{\sigma})=f_\theta(\mathbf{s})$ and is trained on simulated signal-position pairs using the Gaussian negative log-likelihood $\mathcal{L}_{\mathrm{NLL}}=2^{-1}[(\mathbf{x}-\boldsymbol{\mu})^\top\mathbf{\Sigma}^{-1}(\mathbf{x}-\boldsymbol{\mu})+\log\det\mathbf{\Sigma}+3\log(2\pi)]$. The posterior mean provides the position estimate, while $\boldsymbol{\sigma}$ represents coordinate-wise uncertainty conditioned on the assumed simulation model.\looseness=-1


\textbf{The training data} are generated using the \gls{atdm} \citep{owe2026atdm}. Interaction positions are sampled uniformly throughout the detector volume, and the resulting signals are processed through the simulated readout electronics. Events are filtered by requiring an energy deposition of at least $400$~\gls{adc} channels in one anode signal. During training, Gaussian noise $\mathcal{N}(0,5.606)$ is added to the simulated waveforms, with the standard deviation estimated from electronic noise in the experimental data. Training details are provided in Appendix~\ref{app:model_architecture}.

\begin{wrapfigure}[18]{R}{0.45\linewidth}
    \centering
    \includegraphics[width=\linewidth]{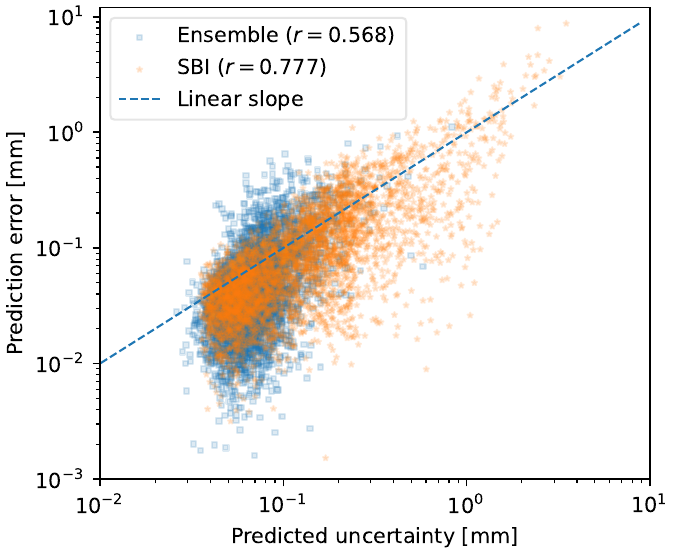}
    \caption{Relationship between predicted uncertainty and positioning error on the synthetic test set for the deep ensemble and SBI. Each point represents an event; the dashed line indicates the $1{:}1$ relationship, while $r$ denotes the Pearson correlation coefficient.}
    \label{fig:uncertainty-correlation}
\end{wrapfigure}

\textbf{Evaluation.~~}
Both approaches are first evaluated on the same synthetic test set using positioning error and its spatial distribution across the detector volume. Their uncertainty estimates are additionally compared with the observed positioning error. The trained models are then applied to real detector data, where positioning resolution is evaluated using slit-beam measurements in the $x$ and $z$ directions. Diagonal slit-beam and full-illumination measurements provide qualitative assessments of localized and global reconstruction behavior, respectively.

\section{Results}
\label{sec:results}

\begin{figure}[htbp]
    \centering
        \includegraphics[width=0.4\textwidth]{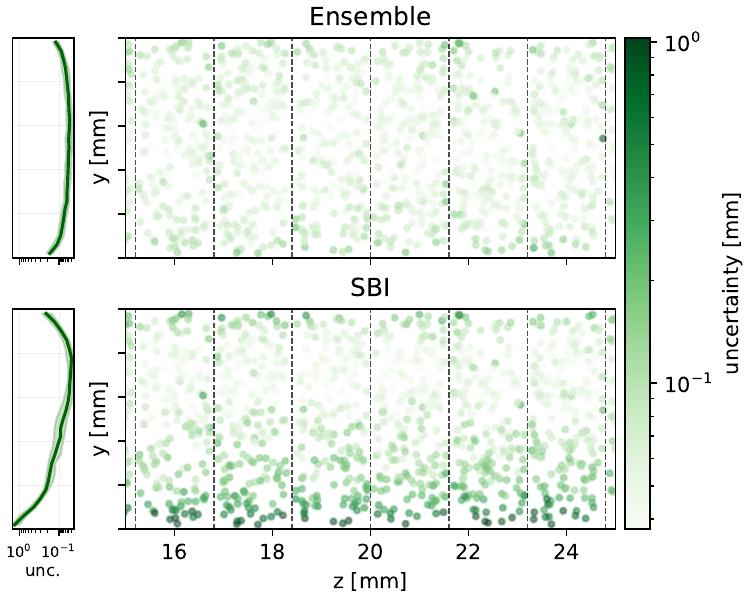}
    \hspace{10mm}
        \includegraphics[width=0.4\textwidth]{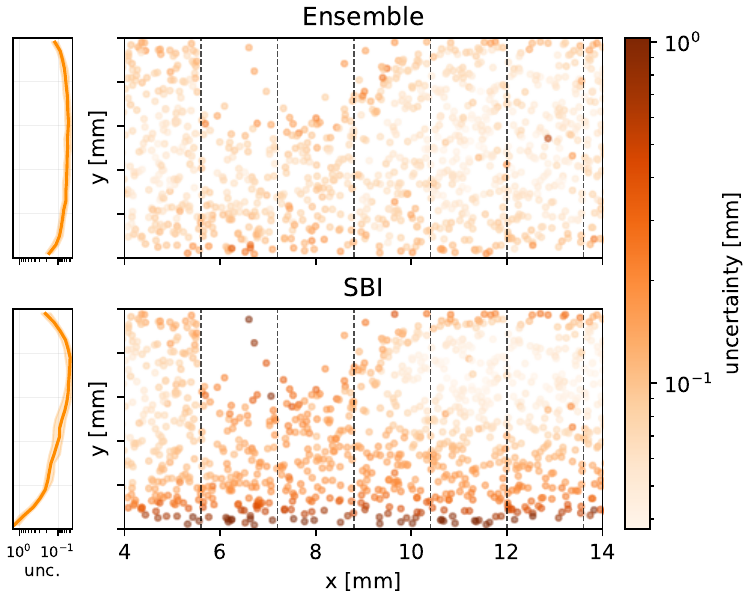}
    \caption{Uncertainty estimates for synthetic-data events under full illumination across the anode ($x$--$y$) and cathode ($z$--$y$) dimensions. \emph{Left: } Uncertainty across anodes in the $x$--$y$ plane; \emph{Right: } Uncertainty across cathodes in the $z$--$y$ plane. \emph{Top:} ensembles; \emph{Bottom:} \gls{sbi}. The corresponding full-range uncertainty estimates are shown in Figure~\ref{fig:xy-uncertainty-synthetic-full}~and~\ref{fig:zy-uncertainty-synthetic-full}.}
    \label{fig:uncertainty-analysis-synthetic}
\end{figure}

\textbf{Synthetic-data uncertainty.~~}
Both methods were evaluated on the synthetic test set, where the true interaction position is known. Fig.~\ref{fig:uncertainty-analysis-synthetic} shows the predicted uncertainty, while Fig.~\ref{fig:uncertainty-correlation} shows the relationship with the corresponding positioning error. Both methods exhibit a positive correlation between predicted uncertainty and positioning error, indicating that uncertainty relates to reliability. This relationship is substantially stronger and more clearly structured for \gls{sbi}. In particular, the \gls{sbi} uncertainty increases toward the lower $y$ boundary, where the available signal information becomes progressively less informative, and position reconstruction is more ambiguous. The ensemble uncertainty shows a considerably weaker spatial structure.

These results demonstrate that both approaches provide useful information about reconstruction reliability, but capture different aspects of uncertainty. The stronger correspondence between \gls{sbi} uncertainty and positioning error suggests that its posterior uncertainty more directly reflects the information content of the observed waveform.

\textbf{Position resolution.~~}
Position resolution is measured using slit-beam illumination, for which the reconstructed interaction positions are expected to follow a Gaussian-like distribution. We quantify the resolution using the \gls{fwhm} of the resulting distribution.

\begin{figure}[htbp]
\centering
\includegraphics[width=0.49\linewidth]{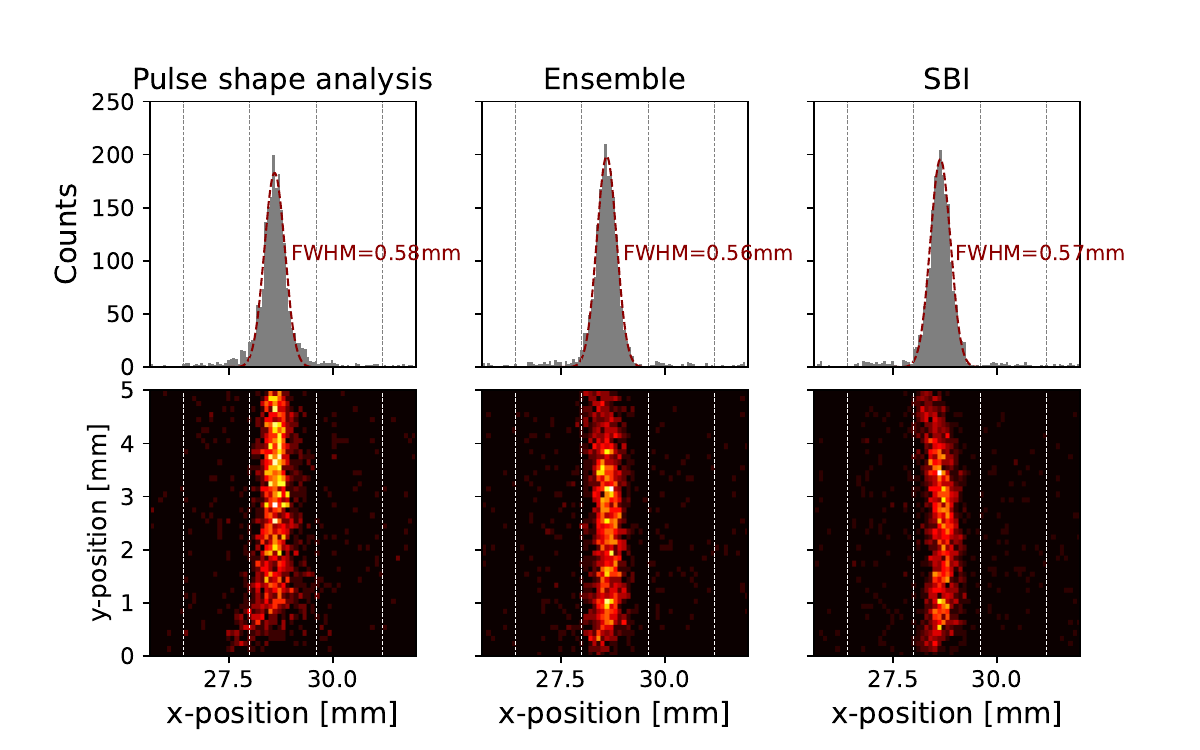}
\hfill
\includegraphics[width=0.49\linewidth]{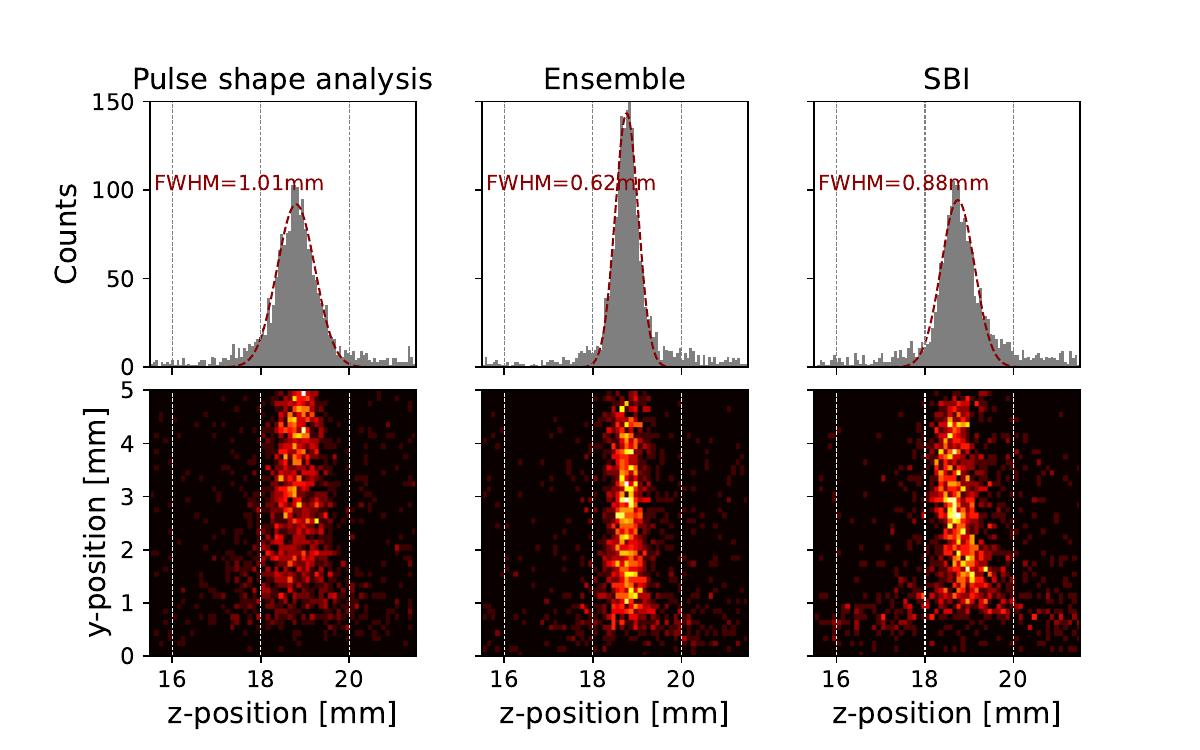}
\caption{Reconstruction resolution for the (a) $x$- and (b) $z$-dimensions using conventional pulse-shape analysis, an ensemble of point-estimate models, and \gls{sbi}. \emph{Left:} pulse-shape analysis; \emph{Center:} ensemble; \emph{Right:} \gls{sbi}. The 1D histograms show the distribution of reconstructed positions, while the 2D histograms show the same reconstructions against the depth of illumination ($y$-position).}
\label{fig:xandz-beam-analysis}
\end{figure}

In the $x$-dimension, all three methods achieve comparable resolution (\gls{fwhm} = 0.58, 0.56, and 0.57~mm for \gls{psa}, ensemble, and \gls{sbi}, respectively), reflecting the strong constraint provided by the drift-strip geometry, as seen in Figure~\ref{fig:xandz-beam-analysis} a. The $z$-dimension, which relies on more subtle pulse-shape information, shows a larger separation: the ensemble achieves the best resolution (0.62~mm), improving substantially over \gls{psa} (1.01~mm), while \gls{sbi} achieves 0.88~mm (Figure~\ref{fig:xandz-beam-analysis} b). The 2D distributions show that reconstruction becomes less accurate near the anodes, consistent with the reduced signal information in this region.

\begin{figure}[htbp]
\centering

\includegraphics[width=0.4\textwidth]{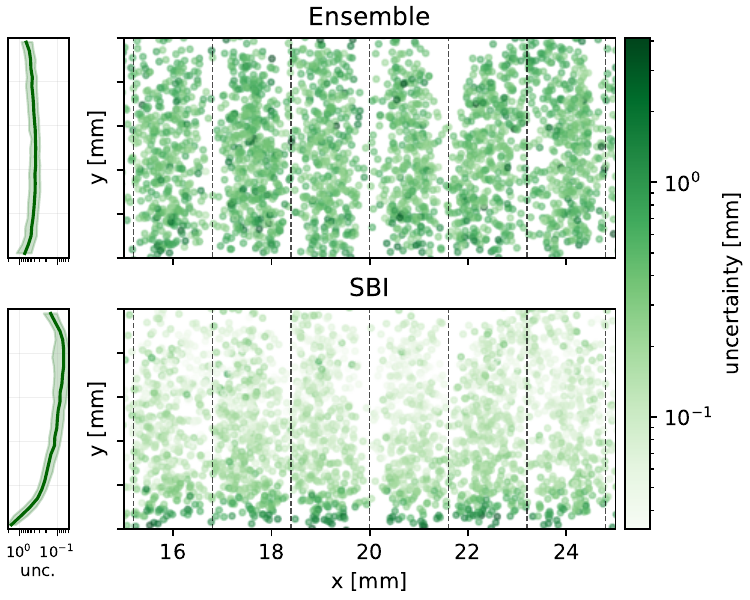}
\hspace{10mm}
\includegraphics[width=0.4\textwidth]{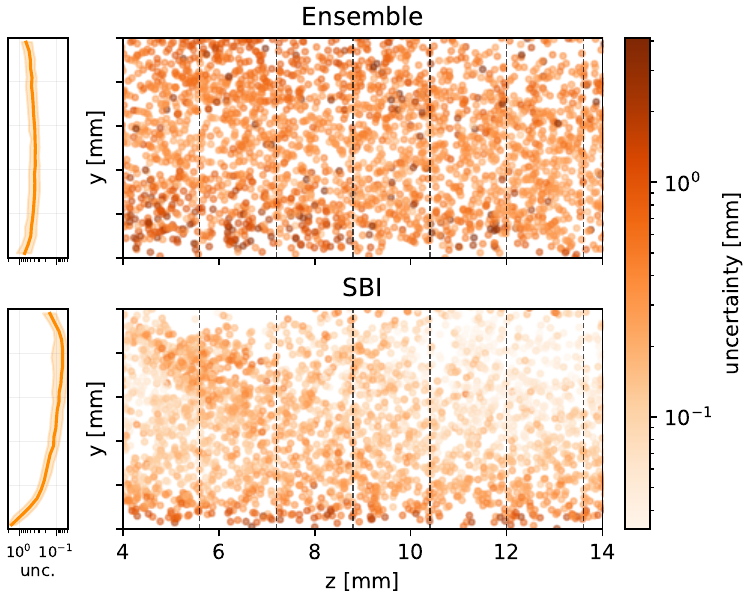}
\hfill

\caption{Uncertainty estimates for real-data events under full illumination across the anode ($x$--$y$) and cathode ($z$--$y$) dimensions. \emph{Left:} Uncertainty across anodes in the $x$--$y$ plane; \emph{Right:} Uncertainty across cathodes in the $z$--$y$ plane. \emph{Top:} Ensemble; \emph{Bottom:} \gls{sbi}. Figure~\ref{fig:xy-uncertainty-real-full}~and~\ref{fig:zy-uncertainty-real-full}}.
\label{fig:uncertainty-analysis-real}
\end{figure}

\textbf{Transfer of uncertainty to real data.~~}
The spatial structure observed on synthetic data also appears in the real measurements. Figure~\ref{fig:uncertainty-analysis-real} shows uncertainty estimates for a full-illumination measurement. For \gls{sbi}, uncertainty increases toward low $y$ in both $x$ and $z$, consistent with the reduced signal information in this region. The ensemble uncertainty shows substantially less spatial structure, with high- and low-uncertainty events distributed more uniformly across the detector.


This difference is consistent with the quantities represented by the two methods. The ensemble measures disagreement between independently trained point estimators and therefore primarily reflects epistemic uncertainty, whereas the \gls{sbi} posterior represents uncertainty in the inferred position conditioned on the observed waveform under the assumed simulation model. The agreement between the synthetic and real-data patterns suggests that the \gls{sbi} uncertainty captures a physically meaningful limitation of the reconstruction.




\begin{figure}[htbp]
    \centering
    \includegraphics[width=\textwidth]{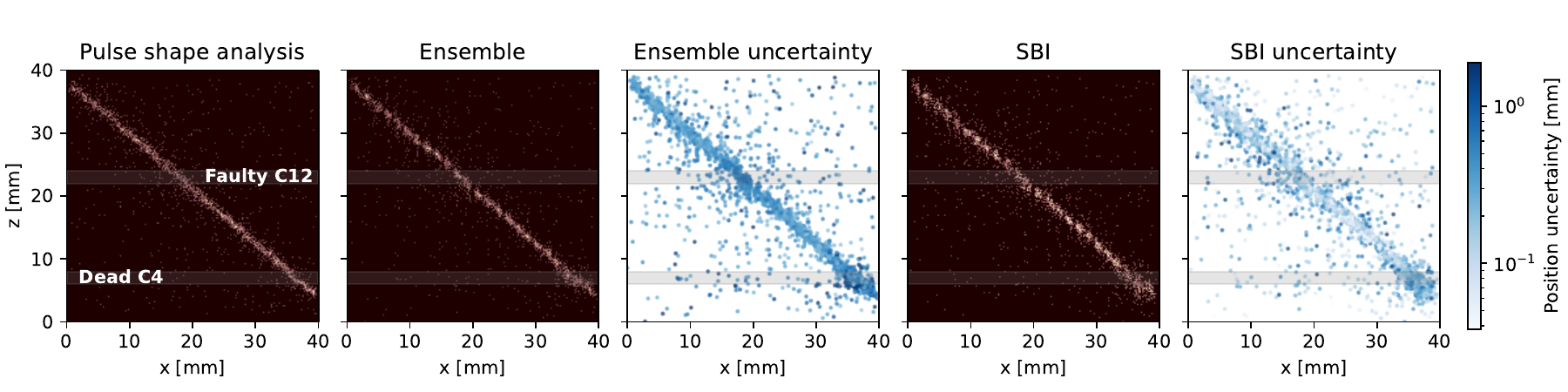}
    \caption{Diagonal slit-beam reconstruction for conventional pulse-shape analysis (first from left), ensemble and its uncertainty (second and third), and \gls{sbi} and its uncertainty (fourth and fifth), viewed in the $x$--$z$ plane. The beam crosses both anode and cathode strips, including faulty cathode \textsc{C12} and dead cathode \textsc{C4}.}
    \label{fig:diagonal-beam-analysis}
\end{figure}

\textbf{Robustness to detector anomalies.~~}
The diagonal slit-beam measurement crosses both Faulty \textsc{C12} and Dead \textsc{C4}, providing a test of detector anomalies that are absent or incompletely represented in the training data. Figure~\ref{fig:diagonal-beam-analysis} shows that \gls{psa} maintains a relatively tight reconstruction through both defect regions, while the ensemble exhibits a modest increase in spread. \gls{sbi} shows the strongest increase in uncertainty around the defects, accompanied by increased reconstruction spread.

This response is particularly notable for \textsc{C4}, whose detector behavior is not fully represented in the \gls{atdm}. The localized increase in \gls{sbi} uncertainty therefore provides an indication of a discrepancy between the simulated and experimental detector response. While this reduces reconstruction accuracy in the affected region, the corresponding increase in posterior uncertainty provides information about when the model's prediction should be treated with caution.


\section{Discussion and Conclusion}
\label{sec:discussion}

The results demonstrate that uncertainty can provide information beyond point-estimate accuracy. On synthetic data, uncertainty from both approaches correlates with positioning error, with the \gls{sbi} posterior showing the stronger relationship. This behavior transfers to real measurements, where the \gls{sbi} uncertainty increases in regions known to provide less positional information and responds to detector anomalies that are absent or incompletely represented in the \gls{atdm}. In particular, the increase around the defective \textsc{C4} electrode suggests that posterior uncertainty can act as a diagnostic signal for discrepancies between the simulated and physical detector.

The current \gls{sbi} model is nevertheless limited to a single Gaussian posterior with diagonal covariance. This prevents it from representing correlations between spatial coordinates or multimodal position distributions, which may become important for ambiguous signals and multiple interactions. Future work will therefore investigate full covariance modelling and more expressive posterior parameterizations, including deep ensembles of posterior estimators \citep{lakshminarayanan_simple_2017}. For events containing multiple interactions, mixtures of Gaussians could provide a natural extension by representing multiple plausible interaction positions. More flexible neural density estimators, such as normalizing flows, could further capture non-Gaussian and multimodal posteriors without prescribing their form.

Overall, the ensemble provides improved point-estimate resolution, particularly in the $z$-dimension, while \gls{sbi} provides richer information about reconstruction reliability and exposes discrepancies between simulation and experiment. Rather than viewing these approaches as competing solutions, their complementary strengths motivate combining accurate ensemble-based reconstruction with expressive posterior modelling. Such uncertainty-aware reconstruction could provide both improved localization and a practical feedback signal for refining the \gls{atdm} and extending the method to multiple energies and interaction types.

\clearpage
\subsection*{Acknowledgments}

This work was supported by the \textit{i-RASE} project, funded by the European Union under the Horizon Europe EIC Pathfinder Open programme (Grant Agreement No. 101130550).


\bibliographystyle{plainnat}
\bibliography{references}

\clearpage

\appendix

\section{Data generation with the Advanced Theoretical Detector Model}
\label{sec:atdm}

Training data are generated using the Advanced Theoretical Detector Model (ATDM) \citep{owe2026atdm}, a physics-based simulation framework that models the detector response from an interaction position to the digitized detector waveform. The underlying charge-transport and signal-induction formulation follows the approach described in \citet{prettyman_theoretical_1999,kolstein_simulation_2014}.

The simulation can be divided into two main stages. First, the charge transport and charge induction in the detector are simulated to obtain the time-dependent charge-induction efficiency (CIE). Second, the resulting CIE signals are passed through a model of the experimental readout electronics to obtain waveforms in ADC units.

\paragraph{Charge transport and charge induction}

The charge-transport and signal-induction model is based on the electrostatic field equations and the Shockley--Ramo theorem. The electrostatic potential $\phi$ is obtained by solving the Laplace equation within the detector geometry,
\begin{equation}
\nabla^2 \phi = 0,
\label{eq:laplace}
\end{equation}
with boundary conditions determined by the applied electrode voltages.

The induced signal on each electrode is described according to the Shockley--Ramo theorem through a corresponding weighting potential $\psi_k$. The weighting potential is obtained by solving the Laplace equation for electrode $k$ with that electrode set to unit potential and the remaining electrodes set to zero. In short, $\phi$ determines how charge carriers move through the detector, while $\psi_k$ determines how their motion is observed by each electrode.

For a given interaction, the resulting carrier transport and charge induction are described by
\begin{equation}
\frac{dn^+}{dt}
=
\mu_e \nabla\phi \cdot \nabla\psi_k
+ \nabla \cdot (\mu_e n^+ \nabla \phi)
+ \nabla \cdot (D_e \nabla n^+)
- \frac{n^+}{\tau_n},
\label{eq:electron_concentration_adj}
\end{equation}
and
\begin{equation}
\frac{dp^+}{dt}
=
\mu_h \nabla\phi \cdot \nabla\psi_k
- \nabla \cdot (\mu_h p^+ \nabla \phi)
+ \nabla \cdot (D_h \nabla p^+)
- \frac{p^+}{\tau_h},
\label{eq:hole_concentration_adj}
\end{equation}
where $n^+$ and $p^+$ denote the electron and hole concentrations, respectively, $\mu_e$ and $\mu_h$ are the corresponding carrier mobilities, and $D_e$ and $D_h$ are their diffusion coefficients. The quantities $\tau_n$ and $\tau_h$ describe the carrier lifetimes. The electrostatic potential is denoted by $\phi$, while $\psi_k$ is the weighting potential associated with electrode $k$. Thus, the equations account for the effects of drift, diffusion, carrier trapping, and charge induction on the electrode.

Solving these equations for the three spatial coordinates and time produces the time-dependent charge-induction efficiency,
\begin{equation}
\text{CIE}_k = \text{CIE}_k(x,y,z,t),
\label{eq:cie}
\end{equation}
which describes the induced response on electrode $k$ for an interaction at a given position.

The resulting four-dimensional solutions are used as look-up tables during data generation. Since calculating the CIE over the entire detector volume for all $69$ electrodes is computationally expensive, the detector volume is instead decomposed into a set of representative sub-volumes. These include anode-, drift-, and cathode-centered configurations, with central and edge variants, as well as a configuration accounting for an inactive cathode in the detector used for experimental validation. The resulting CIE solutions are translated across the detector geometry, allowing the response at arbitrary interaction positions to be obtained efficiently without repeatedly solving the underlying PDEs.

For an interaction depositing an energy $E$, the normalized CIE is converted to charge according to
\begin{equation}
    Q_k(t)
    =
    \text{CIE}_k(t)
    \frac{qE}{E_{\text{ion}}},
    \label{eq:cie_to_charge}
\end{equation}
where $q$ is the elementary charge and $E_{\text{ion}}$ is the mean ionization energy of the detector material. For the simulations used in this work, $E = 662\,$ keV

\paragraph{Electronics simulation}

The induced charge is subsequently converted into a digitized waveform through a model of the experimental readout electronics. The main stages are the charge-sensitive preamplifier, amplification, analog-to-digital conversion, and trigger alignment.

The charge-sensitive preamplifier is modeled as a temporal filtering operation,
\begin{equation}
    V_k(t)
    =
    \mathcal{H}_{\text{CSP}}\{Q_k(t)\},
    \label{eq:csp}
\end{equation}
where $\mathcal{H}_{\text{CSP}}$ represents the response of the charge-sensitive preamplifier and accounts for the channel-dependent input capacitance and preamplifier response.

The resulting analog signal is amplified and converted to ADC units. For an ADC with $b$ bits and voltage range $[V_{\min}, V_{\max}]$, the voltage corresponding to one ADC count is
\begin{equation}
    V_{\text{LSB}}
    =
    \frac{V_{\max}-V_{\min}}{2^b}.
    \label{eq:adc_lsb}
\end{equation}

The digitized signal is then given by
\begin{equation}
    S_k(t)
    =
    \frac{G V_k(t)}{V_{\text{LSB}}},
    \label{eq:adc}
\end{equation}
where $G$ is the amplifier gain. Finally, the simulated waveforms are aligned according to the trigger convention of the experimental acquisition system.

The resulting output consists of $69$ channels of digitized waveforms with the same format as the experimental data. The complete simulation pipeline can therefore be summarized as
\begin{equation}
    (\mathbf{x},E)
    \rightarrow
    \text{CIE}(x,y,z,t)
    \rightarrow
    Q(t)
    \rightarrow
    V(t)
    \rightarrow
    S_{\text{ADC}}(t).
    \label{eq:simulation_pipeline}
\end{equation}

These simulated signal-position pairs are subsequently used for training and evaluating the neural-network and \gls{sbi} approaches.

\section{Detailed model architecture}
\label{app:model_architecture}

The complete architecture is composed of a waveform encoder, a feature encoder, and task-specific prediction heads. The main architecture uses a base channel width of $4$, a waveform embedding dimension of $16$, and a final latent dimension of $32$.

\paragraph{Waveform encoder.}
Each of the $69$ input channels is processed independently by the same one-dimensional convolutional encoder. Given a waveform $\mathbf{x}\in\mathbb{R}^{500}$, an initial convolution with kernel size $7$ maps the single input channel to $4$ feature channels. This is followed by a residual block consisting of two convolutional layers with kernel size $3$. The number of channels is then increased to $8$ while downsampling the temporal dimension by a factor of two. A second residual block is applied, followed by another convolution that increases the number of channels to $16$ and reduces the temporal dimension from $250$ to $125$. A final residual block is followed by a fully connected layer mapping the resulting $16\times125$ representation to a $16$-dimensional waveform embedding.

The residual blocks consist of two convolutional layers with kernel size $3$ and constant channel dimension. GELU activations are applied after the first convolution and after the residual addition.

\paragraph{Feature encoder.}
The same waveform encoder is applied independently to all $69$ channels. The resulting $69$ embeddings, each of dimension $16$, are concatenated to form a $69\times16=1104$-dimensional representation. A fully connected layer then maps this representation to the final $32$-dimensional latent vector $\mathbf{z} \in \mathbb{R}^{32}$.

\paragraph{Autoencoder pretraining.}
During pretraining, the latent representation is passed to a symmetric feature decoder. A fully connected layer first maps the $32$-dimensional latent vector to $69$ waveform embeddings of dimension $16$. Each embedding is then decoded independently by a shared one-dimensional convolutional decoder. The decoder first maps the embedding to a $16\times125$ representation and progressively increases the temporal resolution from $125$ to $250$ and from $250$ to $500$. The channel dimension is simultaneously reduced from $16$ to $8$, then to $4$, and finally to a single output channel. Linear interpolation is used for temporal upsampling, followed by convolutional and residual blocks.

The autoencoder is trained to reconstruct the original normalized signal. During training, the input to the encoder is augmented with additive channel-dependent noise, while the reconstruction target is the corresponding unaugmented signal. This results in a denoising-style pretraining objective, encouraging the latent representation to capture the underlying waveform structure rather than the particular realization of the noise.

\paragraph{Position-reconstruction head.}
Following autoencoder pretraining, the decoder is removed and the pretrained feature encoder is used as the representation network. A three-layer MLP maps the $32$-dimensional latent representation to the three spatial coordinates. The first two layers contain $64$ hidden units and use GELU activations, while the final layer contains three units corresponding to the normalized $x$, $y$, and $z$ coordinates.

\paragraph{\gls{sbi} head.}
For the \gls{sbi} experiments, the same pretrained feature encoder is used. The position head is replaced by an MLP with the same two hidden layers of $64$ units and GELU activations, but with six output values. These parameterize the posterior distribution used by the \gls{sbi} model, with three outputs describing the location parameters and three outputs describing the corresponding uncertainty parameters.

\paragraph{Training details.}
All models are trained using a batch size of $64$ for $2000$ epochs with the \textsc{AdamW} optimizer and a learning rate of $10^{-4}$. The dataset contains $42{,}186$ simulated events, which are split into training, validation, and test sets using an $80/10/10$ split. The dataset split is fixed using a random seed of $42$, while the data shuffling uses a separate fixed seed. Hyperparameters were selected empirically by evaluating a range of configurations on the validation set, and the configuration yielding the best validation performance was used for the final experiments. 

The implementation uses \texttt{JAX} \citep{jax2018github} and \texttt{Equinox} \citep{kidger2021equinox}, allowing the shared waveform encoder to be applied to the individual detector channels using vectorized operations.

\paragraph{Computational resources.}
All experiments were performed on a High-Performace Computing cluster using NVIDIA A100 PCIe GPUs with 40~GB of GPU memory. A single model training run required approximately 3 hours of GPU time. The total computational cost of the experiments reported in this work was approximately 50 GPU-hours across all training runs.

\section{Full-Range Reconstruction Results}
\label{app:appendix-results}

This appendix provides the full-range versions of the reconstruction results shown in the main paper. The main-text figures use zoomed views to improve readability.

\begin{figure}[htbp]
    \centering
    \includegraphics[width=0.8\textwidth]{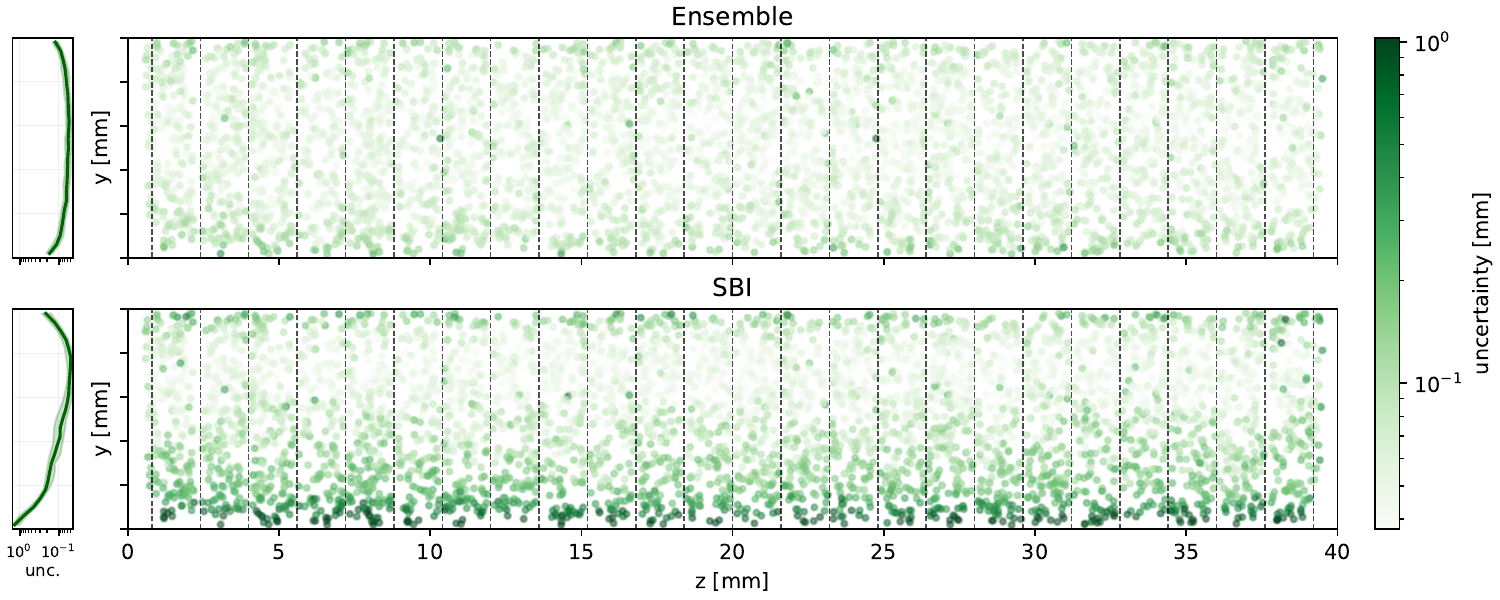}
    \caption{Uncertainty estimates for synthetic-data events under full illumination across the anode dimension. Uncertainty is shown in the $x$--$y$ plane, with the ensemble at the top and \gls{sbi} at the bottom.}
    \label{fig:xy-uncertainty-synthetic-full}
\end{figure}

\begin{figure}[htbp]
    \centering
    \includegraphics[width=0.8\textwidth]{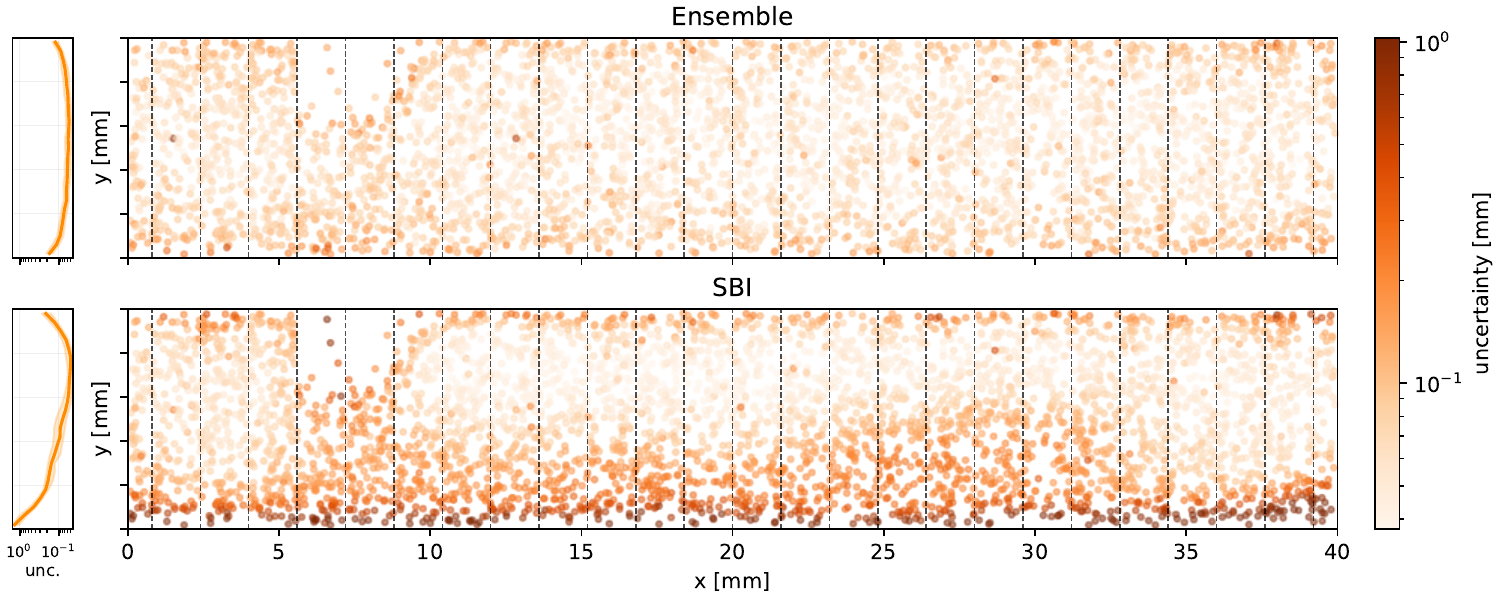}
    \caption{Uncertainty estimates for synthetic-data events under full illumination across the cathode dimension. Uncertainty is shown in the $z$--$y$ plane, with the ensemble at the top and \gls{sbi} at the bottom.}
    \label{fig:zy-uncertainty-synthetic-full}
\end{figure}

\begin{figure}[htbp]
    \centering
    \includegraphics[width=0.8\textwidth]{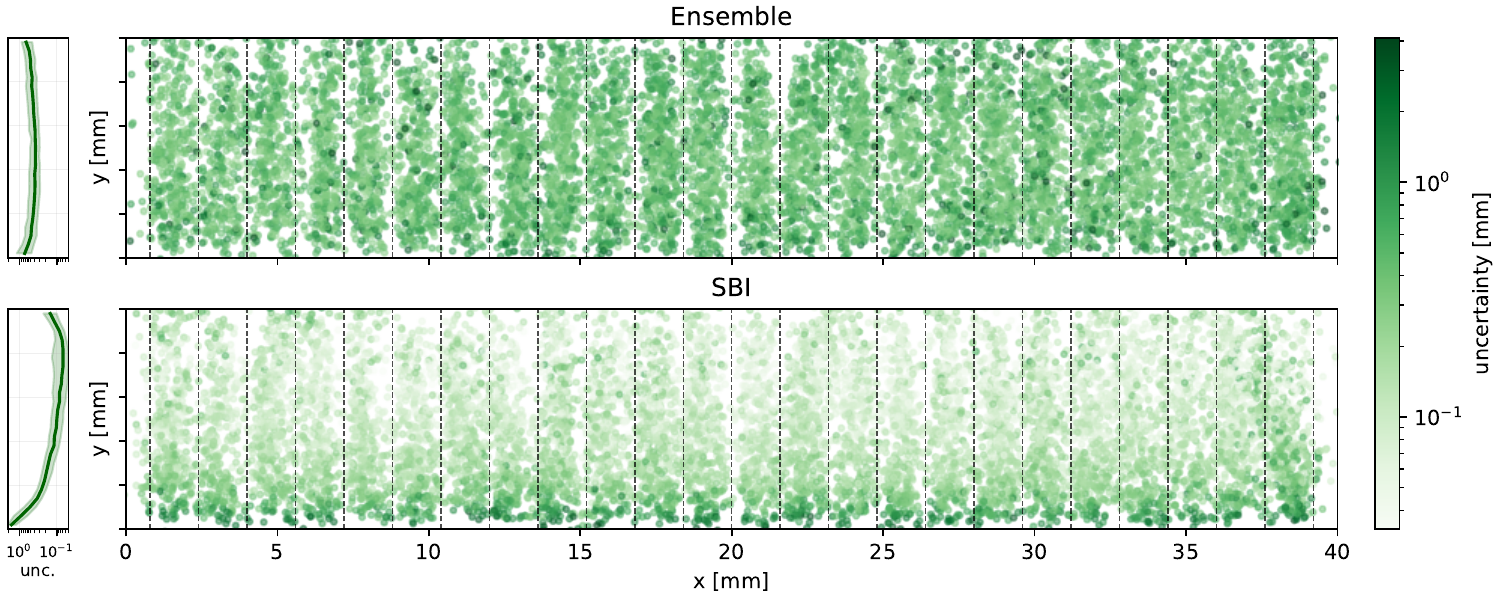}
    \caption{Uncertainty estimates for real-data events under full illumination across the anode dimension. Uncertainty is shown in the $x$--$y$ plane, with the ensemble at the top and \gls{sbi} at the bottom.}
    \label{fig:xy-uncertainty-real-full}
\end{figure}

\begin{figure}[htbp]
    \centering
    \includegraphics[width=0.8\textwidth]{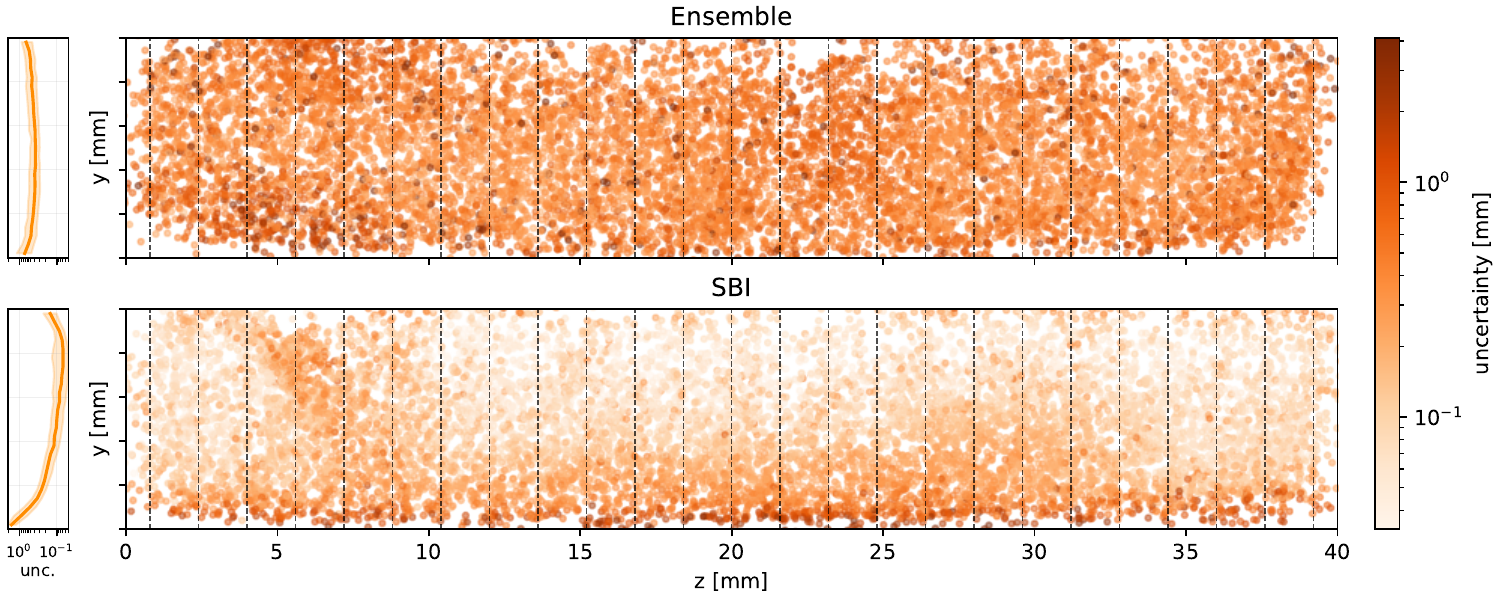}
    \caption{Uncertainty estimates for real-data events under full illumination across the cathode dimension. Uncertainty is shown in the $z$--$y$ plane, with the ensemble at the top and \gls{sbi} at the bottom.}
    \label{fig:zy-uncertainty-real-full}
\end{figure}


\end{document}